# The Effectiveness of Kinematic Constraints on The Accuracy of Trajectory Profile of Human Walking Using PSPB Technique

marwan Qaid Mohammed, muhammad Fahmi Miskon, sari Abdo Ali
Center of Excellence in Robotic and Industrial Automation, Fakulti Kejuruteraan Elektrik, Universiti Teknikal Malaysia Melaka, Hang Tuah Jaya,76100 Durian Tunggal, Melaka, Malaysia.
E-Mail: Marwan.qaid.mohammed@gmail.com

*Abstract*-- **Many methods have been developed in trajectory planning in order to achieve smooth and accurate motion with considering the constraints of kinematics constraints such as angular position, velocity, acceleration, and jerk. The problem of using the combination of n-order polynomials is that there is no ideally match between the segments of trajectory path at the via point in terms of the number of kinematic constraints. It leads to generate undesirable trajectory path at the via point that connects between two segments of the trajectory path. In this paper, we aim to investigate the effect of increasing to higher order polynomial blends on the accuracy of the via points with considering different kinematics constraints. Based on that, the methodology that was used in this paper is based on the polynomial segment with the higher polynomial blend (PSPB). Three techniques implemented which are 4-3-4 PSPB, 5-4-5 PSPB, and 6-5-6 PSPB. Each technique implemented based on applying different kinematic constraints. The three techniques validated using a modeling design in SemiMechanics. According to the methodology, the result analyzed and discussed in terms of angular position, angular velocity, angular acceleration, and angular jerk based on Root Mean Square Error (RMSE) and Average Difference Error (ADE). The result shows that RMSE of angular position for 4-3-4 PSPB-1, 4-3-4 PSPB-2, 5-4-5 PSPB-1, 5-4-5 PSPB-2, 6-5-6 PSPB-1, and 6-5-6 PSPB-2 are 0.4574, 0.0172, 10.9089, 0.1242, 0.6153, and 0.3128 degrees respectively. At the same time, the ADE are 0.0455, 0.0017, 1.0855, 0.0124, 0.0612, and 0.0311 degrees respectively. Thus, the error is increased obviously when there is no ideal match at the via point in terms of a number of kinematic constraints.**

*Index Term*-- **Trajectory Planning, Via Point, Polynomial Segment with Polynomial Blend ( PSPB), Kinematic Constraints, 4-3-4 PSPB, 5-4-5 PSPB, 6-5-6 PSPB, SimMechanics.**

1. INTRODUCTION

Many methods have developed in trajectory planning to achieve a smooth and more accurate motion with considering the constraints of kinematics parameters such as velocity, acceleration, and jerk[1][2]. Moreover, jerk limitation is important for improving the tracking accuracy and speed as well as reducing the manipulator wears[3][4]. Also, the kinematic constraints are used to adapt the algorithm to the specific robotic manipulator and humanoid robots under consideration. This is a feature of paramount importance in all industrial applications, where the velocity, acceleration and jerk values of the humanoid robots and manipulator must be carefully taken into account for many reasons, including safety. Here it has been chosen to set hard constraints on the maximum and minimum values for the velocity, the acceleration and for the jerk at each joint[5][6].

This paper addresses the problem of the effect of different kinematic constraints on the trajectory generation at the via point that is connected between two segments when there is no matched between kinematic constraints of the final point of the first segment and initial point of the second segment. The common four kinematics constraints of the trajectory generation are the position, velocity, acceleration, and jerk.

Trajectory planning is an active field of the research so there is a vast literature treating this issue. The Linear segment with the parabolic blend (LSPB) method used to solve the linear trajectory (constant velocity) to provide a smooth at initial and final position motion[7][8]. Furthermore, LSPB implemented to generate the trajectory of Designing of the Modular Robot Arm because the arm modular has a limited speed and acceleration[9] that can be achieved by LSPB. Also, LSBP trajectory is switched between maximum, minimum, and zero acceleration in order to achieve the optimal time with respecting to the velocity limits. Besides that, Macfarlane and Croft suggested to use cubic or quintic polynomial instead of the parabolic polynomial at blend time in order to achieve a faster and trackable compared to parabolic polynomial[10]. They aimed to solve the infinite jerk at the initial and final points. Additionally, the Cubic polynomial is a common usage in the robotics. It is also considered the lower order polynomial and the easiest one in term of computation that can achieve a smooth motion in term of position and velocity[11]. As well as, The cubic spline used in Cartesian space planning to make either the velocities or the accelerations at the initial and ending moments controllable for the end effector [6]. The cubic spline also implemented to minimize the maximum absolute value of the jerk along the whole trajectory [12]. Furthermore, the cubic polynomial implemented in [13][14] to generate the trajectory of biped robot. To add to that, the B-Splines used for generating a smooth joint trajectory planning for humanoid robots[15]. Whereas, the quintic polynomial implemented for human trajectory generation for lower limb[16][17]. The quintic Spline implemented for 4D trajectory generation for unmanned aerial vehicles[18]. Besides that, fourth and sixth-order polynomials were implemented in [19] to solve the problem of joint-space trajectory generation with a via point. These new polynomials used a single-polynomial function rather than two-polynomial functions that are matched at the via point.

With increasing the demands for methods and techniques of generating a trajectory that can meet the specification of the kinematic parameters of motion, several techniques have developed by many researchers. These techniques developed by combination different polynomial equations with different order degree in order to achieve Integrative process between different order degree polynomials[20][21]. The 434 trajectory used in [22][23][24] to generate the trajectory planning of manipulators Where the 434 means $4^{th} - 3^{rd} - 4^{th}$ order polynomial. The purpose of this technique was to make the interpolation curve double continuous at all points. Moreover, the 445 Trajectory is implemented in [25]. The 445 trajectory combines fifth-order and fourth-order polynomials in order to satisfy the continuity of jerk and give smooth accelerations on all segments of the planned trajectory. Also, two trajectory planning methods for robotic manipulators introduced by Paolo et, they named them as 545 and 5455 trajectories. Both methods implemented based on an interpolation of a sequence of via points using a combination of 4th and 5th order polynomial functions. These techniques allowed obtaining a continuous-jerk trajectory for improved smoothness and minimum excitation of vibration [26][27].

LSPB is a good technique for generating a trajectory with a limited speed and acceleration because LSBP trajectory is switched between maximum, minimum, and zero acceleration in order to achieve the optimal time with respecting to the velocity limits. On other Hand, LSPB trajectory leads to infinite jerk because the second derivative of parabolic polynomial (initial and final point) is constant that leads to discontinuous acceleration at the initial and final point. At the same time, this trajectory can't be used for longer motion time because it is difficult to be tracked. Whereas, the cubic polynomial considers the lower order polynomial and the easiest one in term of computation that can achieve a smooth motion in term of position and velocity. However, cubic polynomial covers four constraints which are initial and final for both position and velocity where there is no consideration for the acceleration and jerk constraints. It leads to generate a constant acceleration that leads to infinite jerk. At the same time, the quintic polynomial used with six constraints which are initial and final of position, velocity, and acceleration. To include the jerk constraint by using quintic polynomial, we have to replace the jerk constraint instead of one of the six constraints that will lead to generating non-smooth trajectory because there is no matched at the via point if the trajectory is built by multi segments. Besides that, using single higher order polynomials such as $6^{th}$ and $7^{th}$ order polynomial[28] leads to generate undesirable acceleration and it takes a longer time to be executed. Additionally, single higher order polynomial (e.g. 7th ,9th , and so on) can produce Runge's phenomenon where The Runge phenomenon illustrates that equidistant polynomial interpolation of the Runge function will cause wild oscillation near the endpoints of the interpolation interval as the order of the interpolation polynomial increases [29][30][31][32][33].

The combination of the different n-order polynomial is better than using a single n-order polynomial because the combination provides more kinematic constraints that it can't be provided by using single polynomial. However, all combined n-order polynomial techniques that are aforementioned such as 434 trajectory, 443 trajectory, 545 trajectory, and 5455 trajectory, they can't be used for generating a long path because the constraints are not matched in connecting points of the two segment. Which mean that it can be used to generate manipulator trajectory but it is not suitable for generating a walking motion trajectory.

This paper is organized as follows: Section 2 presents Methodology; Section 3 presents result and discussion; section 4 presents the conclusion.

## 2. METHODOLOGY

In this part, the effect of different kinematic constraints on the trajectory generation will be illustrated and discussed. Based on the combination of different type order polynomial, the constraints are different from one combination to other according to the type of n-order polynomial. The common four kinematics constraints of the trajectory generation are position, velocity, acceleration, and jerk. Thus, The combination of ($4^{th}$ -$3^{rd}$ -$4^{th}$), ($5^{th}$ -$4^{th}$ -$5^{th}$) and ($6^{th}$ -$5^{th}$ -$6^{th}$) with considering different constraints, we have the ability and proof to figure out the effect of using different constraints on the trajectory.

### 2.1 The Trajectory of 4-3-4 PSPB

The trajectory planning of the 4-3-4 PSPB technique generates based on the stance and swing phases with consideration to the polynomial blend. The gait motion of human divides into two phases based on the biomechanical analysis: (1) stance phase (2) swing phase. Each phase is presented by one full 4-3-4 PSPB trajectory. The 4-3-4 PSPB trajectory has three segments where the first segment is presented by $4^{th}$ order polynomial, the intermediate segment is presented by $3^{rd}$ order polynomial, and the last segment is presented by $4^{th}$ order polynomial as illustrated in Fig1.

There are two cases that have been involved to figure out the effect of applying different constraints on each single order polynomial. The first case, the constraints have been taken for each single $4^{th}$, $3^{rd}$ and $4^{th}$ order polynomial. For differentiating between the two cases we can call the first case that is illustrated in Table I as 4-3-4 PSPB-1 trajectory (initial acceleration constraint of first segment and final acceleration constraint of the last segment).

The second case, the constraints have been taken for each single $4^{th}$, $3^{rd}$, and $4^{th}$ order polynomial. We named the second case as 4-3-4 PSPB-2 trajectory as illustrated in Table II (mid-point position constraint at first segment and last segment).

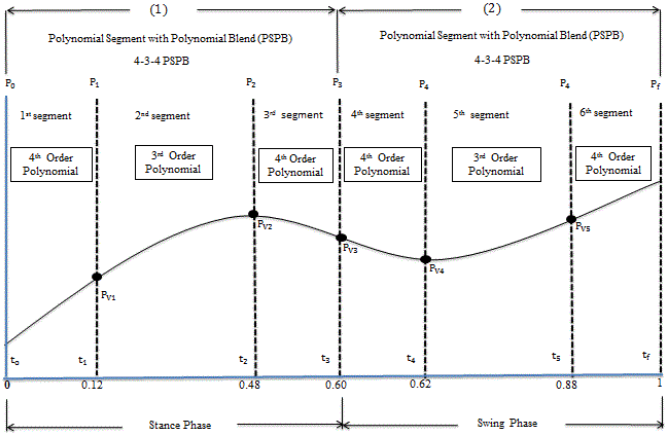
Fig 1: Methodology of the 4-3-4 PSPB technique

## 2.2 The Trajectory of the 5-4-5 PSPB

The trajectory planning of the 5-4-5 PSPB technique generates based on the stance and swing phases with consideration to the polynomial blend. The gait motion of human divides into two phases based on the biomechanical analysis: (1) stance phase (2) swing phase. Each phase is presented by one full 5-4-5 PSPB trajectory. The 5-4-5 PSPB trajectory has three segments where the first segment is presented by $5^{th}$ order polynomial, the intermediate segment is presented by $4^{th}$ order polynomial, and the last segment is presented by $5^{th}$ order polynomial as illustrated in Fig 2.

There are two cases that have been involved to figure out the effect of applying different constraints on each single order polynomial. The first case, the constraints have been taken for each single $5^{th}$, $4^{th}$, and $5^{th}$ order polynomial. For differentiating between the two cases, we named the first case as 5-4-5 PSPB-1 trajectory that is illustrated in Table III (initial acceleration and jerk constraint at of first segment and initial acceleration of the mid segment).

The second case, the constraints have been taken for each single $5^{th}$, $4^{th}$, and $5^{th}$ order polynomial. We called the second case as 5-4-5 PSPB-2 trajectory that is illustrated in Table IV (initial and final acceleration constraint at first and last segment).

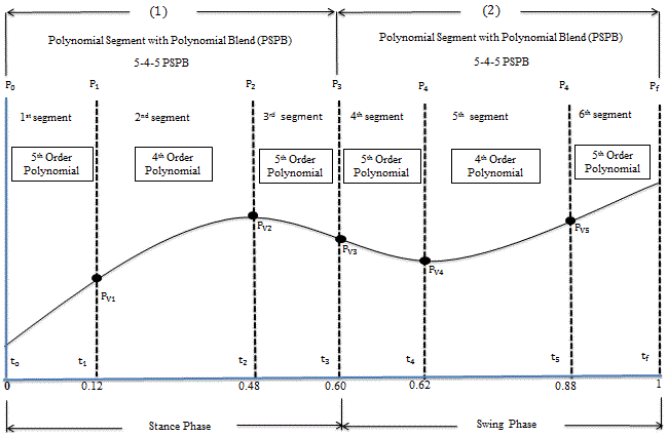
Fig 2: Methodology of the 5-4-5 PSPB technique

## 2.3 The Trajectory of the 6-5-6 PSPB Technique

The trajectory planning of the 6-5-6 PSPB technique generates based on the stance and swing phases with consideration to the polynomial blend. The gait motion of human divides into two phases based on the biomechanical analysis: (1) stance phase (2) swing phase. Each phase is presented by one full 6-5-6 PSPB trajectory. The 6-5-6 PSPB trajectory has three segments where the first segment is presented by $6^{th}$ order polynomial, the intermediate segment is presented by $5^{th}$ order polynomial, and the last segment is presented by $6^{th}$ order polynomial as illustrated in Fig 3.

There are two cases that have been used to figure out the effect of applying different constraints on each single order polynomial. The first case, the constraints have been taken for each single $6^{th}$, $5^{th}$, and $6^{th}$ order polynomial. For differentiate between the two cases, we called the first case as 6-5-6 PSPB-1 trajectory that is illustrated in Table V (initial and final acceleration and jerk constraint at first segment where final acceleration and jerk constraint at last segment).

The second case, the constraints have been taken for each single $6^{th}$, $5^{th}$, and $6^{th}$ order polynomial. We named the second case as 6-5-6 PSPB-2 trajectory that is illustrated in Table VI (initial acceleration and mid-point position constraint at first and final segments).

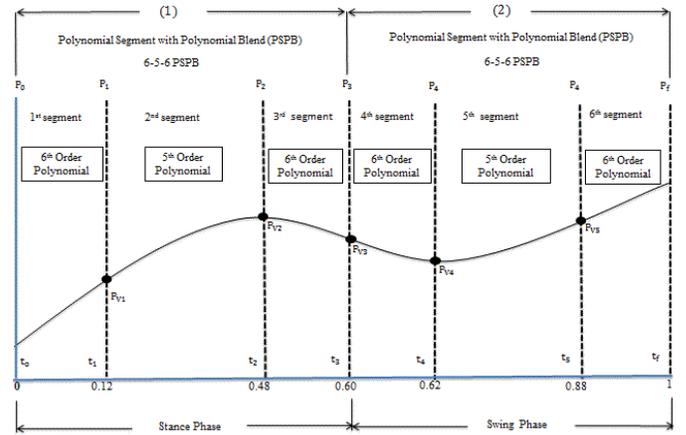
Fig 3: Methodology of the 6-5-6 PSPB technique

Table I
Kinematics constraints of the 4-3-4 PSPB-1

| Order polynomial | Constraints | |
|---|---|---|
| | Initial | Final |
| $4^{th}$ order polynomial | P,V,A | P,V |
| $3^{rd}$ order polynomial | P,V | P,V |
| $4^{th}$ order polynomial | P,V | P,V,A |

Table II
Kinematics constraints of the 4-3-4 PSPB-2

| Order polynomial | Constraints | | |
|---|---|---|---|
| | Initial | Mid-point | Final |
| $4^{th}$ order polynomial | P,V | $P_m$ | P,V |
| $3^{rd}$ order polynomial | P,V | non | P,V |
| $4^{th}$ order polynomial | P,V | $P_m$ | P,V |

Table III
Kinematics constraints of the 5-4-5 PSPB-1

| Order polynomial | Constraints | |
|---|---|---|
| | Initial | Final |
| 5th order polynomial | P,V,A,J | P, V |
| 4th order polynomial | P,V,A | P, V |
| 5th order polynomial | P,V,A | P,V,A |

Table IV
Kinematics constraints of 5-4-5 PSPB-2

| Order polynomial | Constraints | | |
|---|---|---|---|
| | Initial | mid-point | Final |
| 5th order polynomial | P,V,A | non | P,V,A |
| 4rd order polynomial | P,V | $P_m$ | P,V |
| 5th order polynomial | P,V,A | non | P,V,A |

Table V
Kinematics constraints of the 6-5-6 PSPB-1

| Order polynomial | Constraints | |
|---|---|---|
| | Initial | Final |
| 6th order polynomial | P,V,A,J | P,V,A |
| 5rd order polynomial | P,V,A | P,V,A |
| 6th order polynomial | P,V,A | P,V,A,J |

Table VI
Kinematics constraints of the 6-5-6 PSPB-2

| Order polynomial | Constraints | | |
|---|---|---|---|
| | Initial | mid-point | Final |
| 6th order polynomial | P,V,A | $P_m$ | P,V,A |
| 5rd order polynomial | P,V,A | non | P,V,A |
| 6th order polynomial | P,V,A | $P_m$ | P,V,A |

### 2.4 Modeling Design of Simulation

MATLAB is a reliable program that is widely used in an array of fields. In MATLAB, the method is tested to figure out the ability of generating an appropriate trajectory profiles compared with the real human trajectory profile. In addition to that, the role of SemMechanics will be designing and testing the 6-5-6 PSPB trajectory method. SemMechanics is a three-dimensional modeling tool that allows the simulation of multi-body models such as robots. As well, it's interface functions by means of blocks representing bodies (links), joints, motion constraints and forces, and torques. In addition, the various elements are joined together using lines that are representing the transmitted signals from one block to another. Thus, Many researchers used MATLAB Algorithm and SemMechanics to validate the trajectory planning methods of walking motion[34][35]. A dynamic model of leg designed in SimMechanics in order to validate geometrical parameters of the leg[36]. Also, Modeling of human walking developed in SemMechanics to optimize the function of ankle-foot orthosis [37]. Additionally, study of human walking implemented based on SemMechanics [38].

The system that used to validate the PSPB trajectories is illustrated in Fig4. The system consists of trajectory generator, simple controller (PD controller), actuator (joint), and feedback block.

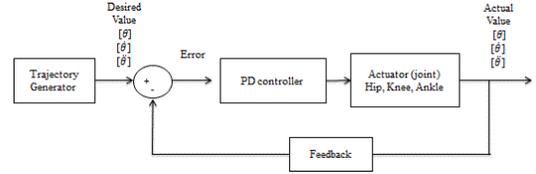

Fig. 4. System of off-line trajectory planning

According to the actuators, two types of actuator are used in leg modeling design: (1) joint actuator and (2) body actuator. The joint actuator is used to actuate the joint by using motion mode. The motion (trajectory planning) sends from the trajectory generator in terms of the angular position, angular velocity, and angular acceleration. As well as, the body actuator is used to implement the torque and the force that are required to move the segment with consideration to the mass and the length of each segment.

In order to validate the result, we designed a modeling of hip link segment in SimMechanic as illustrated in Fig 5. We validated the combination of different PSPB such as 4-3-4 PSPB, 5-4-5 PSPB, and 6-5-6 PSPB with considering different kinematic constraints.

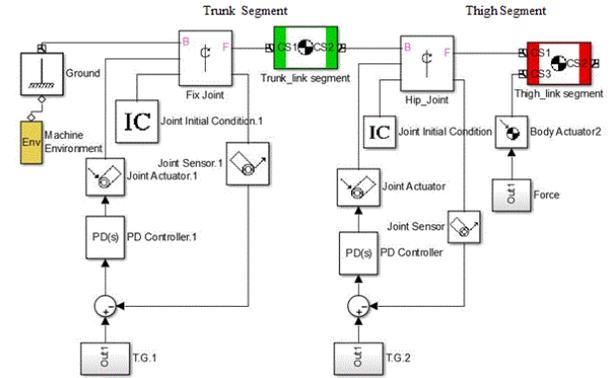

Fig. 5. Modeling design of hip link segment based on SemiMechanics

The leg modeling designed in the sagittal plane. As well, in the leg modeling designing, the mass and the length of each segment are the parameters that are taken in account as stated in Table VII.

Table VII
The mass, length, center of mass of the two body segments [37]

| Body segment | Mass | Length | Centre of mass |
|---|---|---|---|
| Thigh | 15.961 | 0.5287 | 0.3183 |
| Trunk | 17.761 | 0.7050 | 0.2965 |

Based on that, the hip modeling design consists of two segments (trunk and thigh) as illustrated in Fig 6. One joint used in the designing which is hip joint. Body and joint actuators are the two types of actuators that are involved in the hip modeling designing.

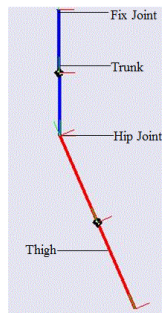

Fig. 6. The Hip Modeling design in SemMechanics

## 3. RESULT AND DISCUSSION

### 3.1 Accuracy analysis of the PSPB techniques

In the part, we will show the result of accuracy test based on analytical techniques such as RMSE and ADE for angular of joint in terms of position, velocity, acceleration, and jerk that will be evaluated and discussed.

#### 3.1.1 The 4-3-4 PSPB technique

The 4-3-4 PSPB technique is used to generate the trajectory based on different constraints. As well as, the 4-3-4 PSPB technique tested and analyzed based on different cases with consideration to the different constraints in each case.

Related to Table VIII, the 4-3-4 PSPB-1 method generated the trajectory profile with higher RMSE and ADE in term of angular position almost 0.4574 degree and 0.0455 degree compared to 4-3-4 PSPB-2. Whereas, less RMSE and ADE are obtained for 4-3-4 PSPB-1 for all the profiles of angular velocity, acceleration, and jerk due to using the acceleration as constraints compared to 4-3-4 PSPB-2. On another hand, the acceleration is the second derivative of the position; it means that $3^{rd}$ will be linear in acceleration profile and constant in jerk profile. This leads to the infinite acceleration that directly affects on the jerk to be infinite spike jerk. Thus, the 4-3-4 PSPB technique is not suitable to generate the trajectory with the smooth transition for acceleration and jerk profile through via Points.

Table VIII
RMSE and ADE based on the angular position, velocity, and acceleration, jerk

| Joint \ Method | | 4-3-4 PSPB-1 | 4-3-4 PSPB-2 |
|---|---|---|---|
| Hip (Pos) | RMSE | 0.4574 | 0.0172 |
| | ADE | 0.0455 | 0.0017 |
| Hip (Vel) | RMSE | 0.2141 | 0.2141 |
| | ADE | 0.0213 | 0.0213 |
| Hip (Accel) | RMSE | 30.2723 | 149.1475 |
| | ADE | 3.0122 | 14.8407 |
| Hip (Jerk) | RMSE | 7297.5864 | 7403.6354 |
| | ADE | 726.1370 | 736.6893 |

#### 3.1.2 The 5-4-5 PSPB technique

Based on Table IX, the 5-4-5 PSPB-1 method generated the trajectory profile with higher RMSE and ADE in term of angular position, velocity, acceleration almost 10.9089, 1.0855 degree, 3.1170, 0.3102 d/s, 442.1972, and 44.0003d/$s^2$ respectively compared to 5-4-5 PSPB-2. Whereas, the less RMSE and ADE obtained for 5-4-5 PSPB-1in term of angular jerk almost 52944.2011 d/$s^3$ due to using the jerk as constraints compared to 5-4-5 PSPB-1. On another hand, the jerk is the third derivative of the position; it means that $4^{th}$ will be linear in jerk profile. This leads to the infinite acceleration that directly affects on generating infinite spike jerk. Thus, 5-4-5 PSPB is not suitable to generate the trajectory with the smooth transition for jerk profile. At the same, it cannot be used to solve the infinite acceleration and jerk through all the segments of the gait human motion.

Table IX
RMSE and ADE based on the angular position, velocity, and acceleration, jerk

| Joint \ Method | | 5-4-5 PSPB-1 | 5-4-5 PSPB-2 |
|---|---|---|---|
| Hip (Pos) | RMSE | 10.9089 | 0.1242 |
| | ADE | 1.0855 | 0.0124 |
| Hip (Vel) | RMSE | 3.1170 | 0.5167 |
| | ADE | 0.3102 | 0.0514 |
| Hip (Accel) | RMSE | 442.1972 | 227.6179 |
| | ADE | 44.0003 | 22.6488 |
| Hip (Jerk) | RMSE | 52944.2011 | 66940.5884 |
| | ADE | 5268.1449 | 6660.8375 |

#### 3.1.3 The 6-5-6 PSPB Technique

Based on Table X, the 6-5-6 PSPB-2 method generated the trajectory profile with higher RMSE and ADE in term of angular, velocity, acceleration, and jerk almost 0.0022, 0.00022 d/s, 41.238, 4.1033d/s2, and 4611.0020, 458.8118d/s3 respectively compared to 6-5-6 PSPB-1. Whereas, the less RMSE and ADE are obtained for 6-5-6 PSPB-1 in term of angular, velocity, acceleration, and jerk due to using the jerk as constraints compared to 6-5-6 PSPB-2.

Table X
RMSE and ADE based on the angular position, velocity, and acceleration, jerk

| Joint \ Method | | 6-5-6 PSPB-1 | 6-5-6 PSPB-2 |
|---|---|---|---|
| Hip (Pos) | RMSE | 0.6153 | 0.3128 |
| | ADE | 0.0612 | 0.0311 |
| Hip (Vel) | RMSE | 0.0022 | 0.0022 |
| | ADE | 0.00022 | 0.00022 |
| Hip (Accel) | RMSE | 21.9663 | 41.238 |
| | ADE | 2.1857 | 4.1033 |
| Hip (Jerk) | RMSE | 3853.6152 | 4611.0020 |
| | ADE | 383.4491 | 458.8118 |

6## 3.2 Simulation and Calculation Result

In the simulation part, we will show the result based on angular of joint in terms of position, velocity, and acceleration that will be evaluated and discussed.

### 3.2.1 Angular Position

Based on Table XI, the 5-4-5 PSPB1 trajectory has discontinuous at the point 0.12s that connects between the first segment and second segment during stance phase. The reason that is the constraints at the end of the first segment and the constraints at the initial of the second segment don't ideally match as indicated by red box in Figure (c) (e.g. at final point of the first segment is P, V, A where P, V, at the initial of the second segment). As well as, during the whole second segment from 0.12 s to 0.48s, there error is higher compared to the reference profile. The reason return to that the constrains at the via point that connects between the first segment and second segment of the stance phase are not correlated to each other in term of number of constraints as indicated by red box in figure (c).

According to the 4-3-4 PSPB technique, the two cases either 4-3-4 PSPB-1 or 4-3-4 PSPB-2 show an accurate trajectory profile compared with reference trajectory profile. That is because that the constraints at the via points are correlated compared to the 5-4-5 PSPB-1. However, 4-3-4 PSPB are a combination of lower order polynomial, it leads to cover least constraints that are not sufficient to generate the trajectory with consideration to the most four effectiveness kinematic constraints on the trajectory human motion.

Related to 6-5-6 PSPB techniques for both two cases, the trajectory profile is more accurate compared to the reference profile as well smooth transition is noticed during the via points compared to the other two techniques (4-3-4 PSPB and 5-4-5 PSPB). The reason that the kinematic constraints at the via point ideally match through the whole segments of the trajectory. Although the 6-5-6 PSPB-2 technique covers angular position, velocity and acceleration through all the via points of the trajectory, the jerk constraint did not take in consideration. In order to involve the jerk constraints at the initial position and the final position of both stance and swing phases of the gait human motion, 6-5-6 PSPB-1 was implemented. The reason that is the crucial situation in the trajectory planning is the initial and final position of the stance and swing phases. At this position of the starting and ending of the gait motion requires a smooth trajectory with consideration to the most effectiveness four kinematic constraints (angular position, velocity, acceleration, and jerk).

### 3.2.2 Angular Velocity

Based on Table XII, the red box indicates that the non-correlation between the constraints will affect on the trajectory profile. At point 0.12s which is the connecting point between the first segment and second segment of the 4-3-4 PSPB trajectory generation method has less effect. The reason is that , either 4-3-4 PSPB-1 or 4-3-4 PSPB-2 has the same constraints at the connecting points between each segment .However, the error that indicates by red box based on both 4-3-4 PSPB-1 and 4-3-4 PSPB-2, it is because of using the constraints acceleration at the initial of the first segment , which means the first segment has (P,V,A) constraints at initial and (P,V) at final point where the second segment has (P,V) at initial and (P,V) at final point. In order to proof that reason, we replace acceleration constraint by the mid- point position based on 4-3-4 PSPB-2. Based on that, we can notice that error terminated and the profile became smooth through via point.

The error can be noticed higher when we used combination of the $5^{th}$ and $4^{th}$ order polynomial. The reason behind that, the constraints are not correlated between at initial and final point of each segment. For 5-4-5 PSPB-1, the first segment has (P, V, A, J) constraints at initial and (P, V) at final point where the second segment has (P, V, A) at initial and (P, V) at final point, which means the error will be more and increased. In order to solve that, we have tried to replace the normal constraints of each single order polynomial as indicated in different text books. We suggested using 5-4-5 PSPB-2 as the trajectory method that can indicate the acceleration at initial point of the first segment and final point of the last segment. The first segment has (P, V, A) constraints at initial and (P, V, A) at final point where the second segment has (P, V,$P_m$) at initial and (P,V) at final point. However, the constraints at the final point of first segment and initial point of the second segment are still not equal. Thus, the error is still available even though the error is reduced compared to first method of the same combination.

Combination of $6^{th}$ and $5^{th}$ order polynomial can be the best in term avoiding the error of non-correlation of the constraints between the initial and final of the first and second segment respectively. Two methods are developed. Each method is different from each other based on the constraints that are involved in trajectory generation. Firstly, 6-5-6 PSPB-2 trajectory generation method has been developed based on the usual constraints that are indicated in text book. The first segment has (P, V, A, $P_m$) constraints at initial and (P,V,A) at final point where the second segment has (P,V,A) at initial and (P,V,A) at final point and the last segment has the same constraint of the first segment. Related to the contribution of constraints, we can see that the constraints are equal in each via point between each segment. However, the jerk constraints is not considering in this method that cannot be used to terminate the infinite spike of the jerk. We rearranged the contribution of the constraints to replace the (J) constraint instead of ($P_m$) constraints. Based on that, the 6-5-6 PSPB-1 is developed and used to generate the trajectory profile.

Based on Table XII in Figure (h), the error of the 6-5-6 PSPB-1 is less at point 0.12s compared to other order polynomial combination as indicated by red box. The reason is that jerk constraints is replace at initial point of the first segment that lead to generate noise. However, the noise is still not much to affect on the trajectory.

Table XI
Comparison Table between the PSPB techniques based on the angular position for hip joint profile

### 4-3-4 PSPB-1          4-3-4 PSPB-1

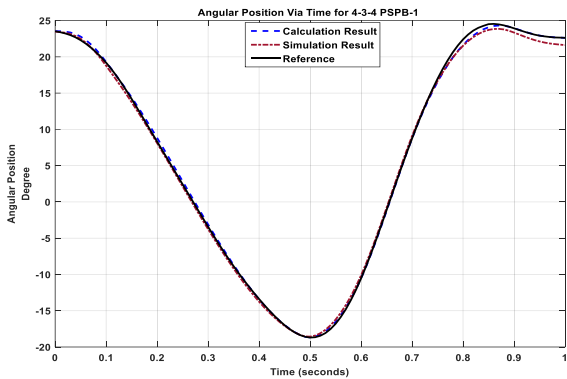 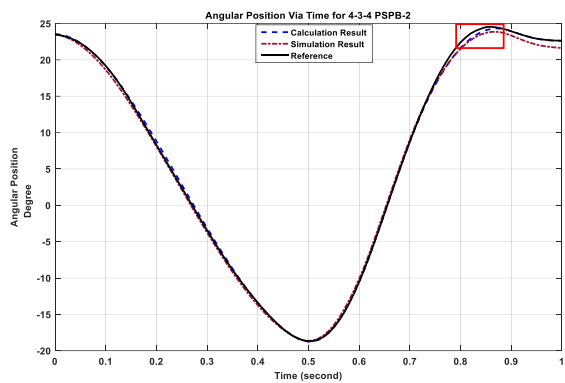

(a)          (b)

### 5-4-5 PSPB-1          5-4-5 PSPB-1

Angular Position

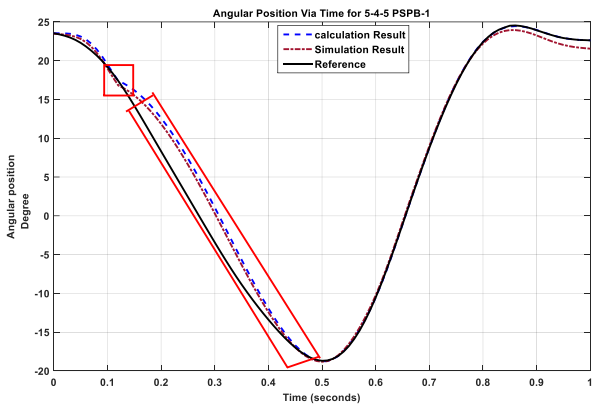 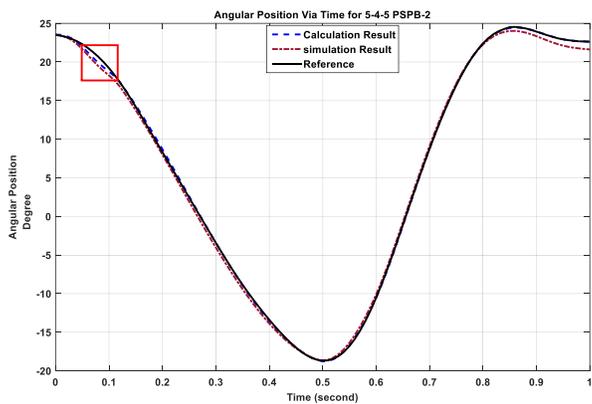

(c)          (d)

### 6-5-6 PSPB-1          6-5-6 PSPB-1

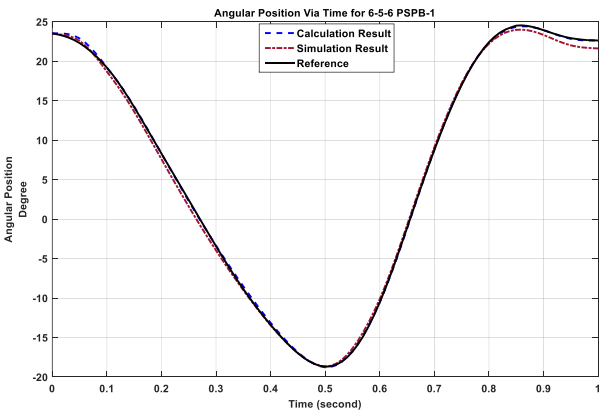 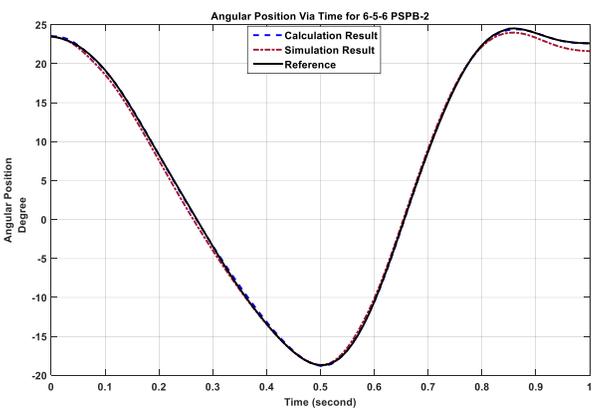

(e)          (f)

Table XII
Comparison Table between the PSPB techniques based on the angular velocity for hip joint profile

4-3-4 PSPB-1              4-3-4 PSPB-1

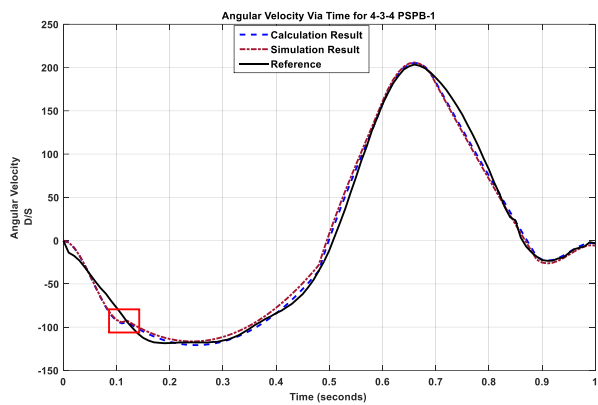
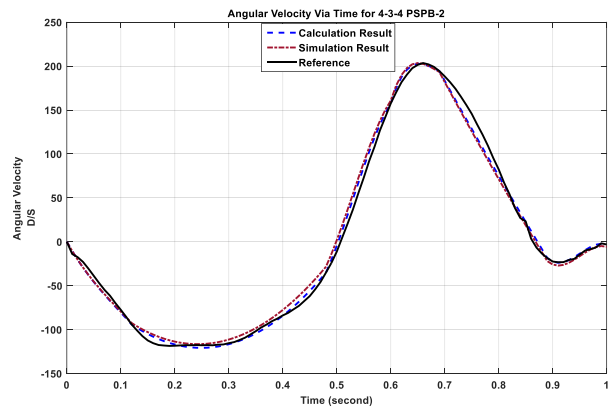

(a)      (b)

5-4-5 PSPB-1              5-4-5 PSPB-1

Angular Velocity

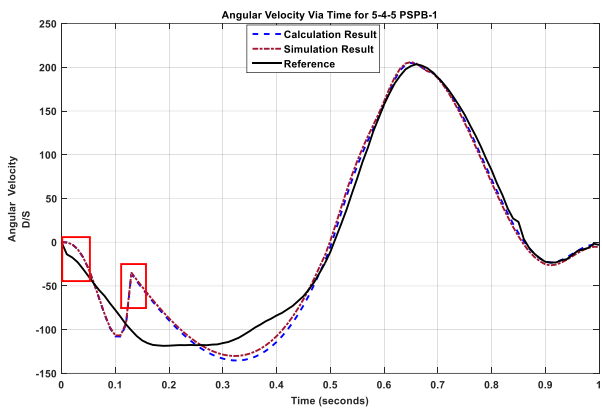
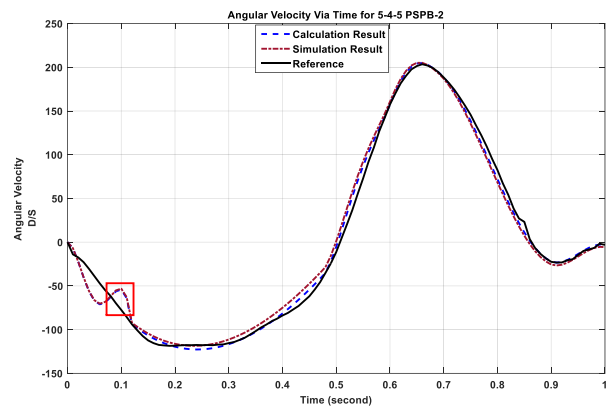

(c)      (d)

6-5-6 PSPB-1              6-5-6 PSPB-1

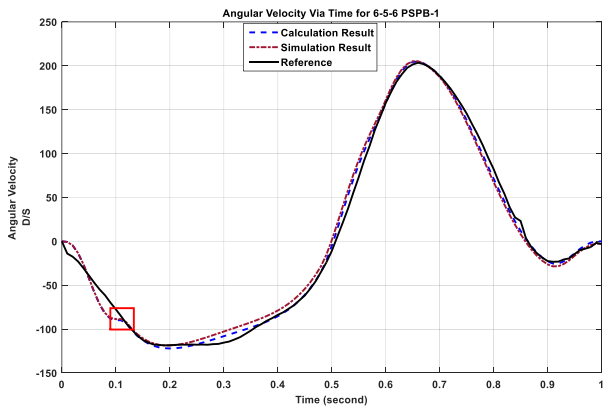
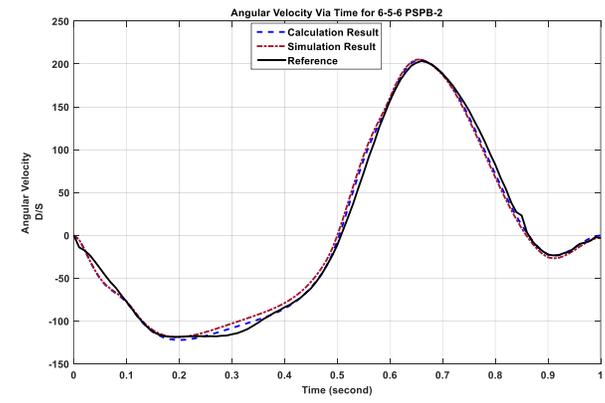

(e)      (f)

Table XIII
Comparison Table between the PSPB techniques based on the angular acceleration for hip joint profile

| | 4-3-4 PSPB-1 | 4-3-4 PSPB-1 |
|---|---|---|
| | 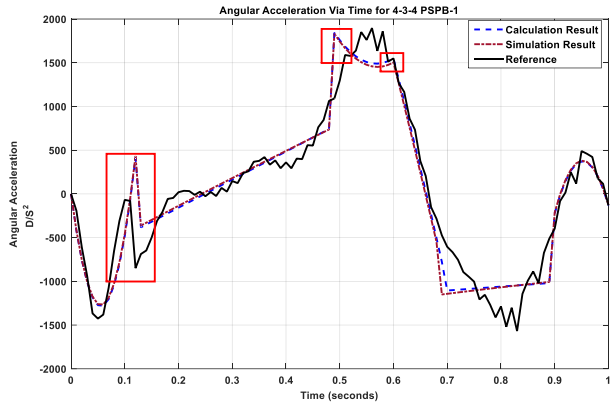 | 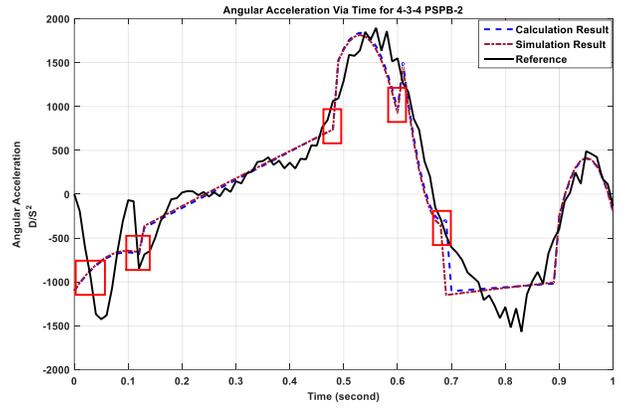 |
| | (a) | (b) |
| | 5-4-5 PSPB-1 | 5-4-5 PSPB-1 |
| Angular Acceleration | 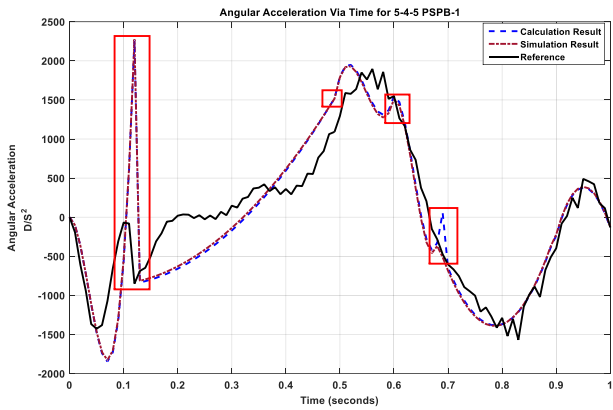 | 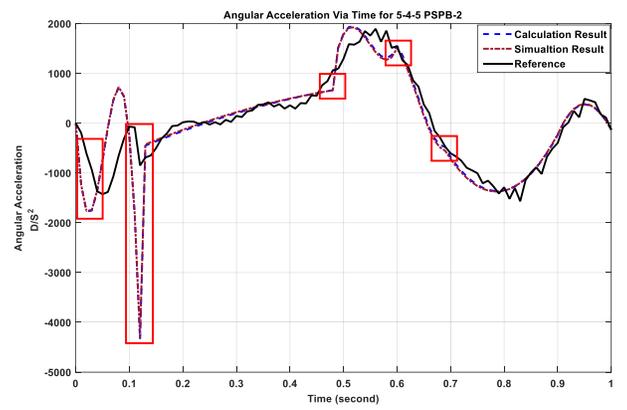 |
| | (c) | (d) |
| | 6-5-6 PSPB-1 | 6-5-6 PSPB-1 |
| | 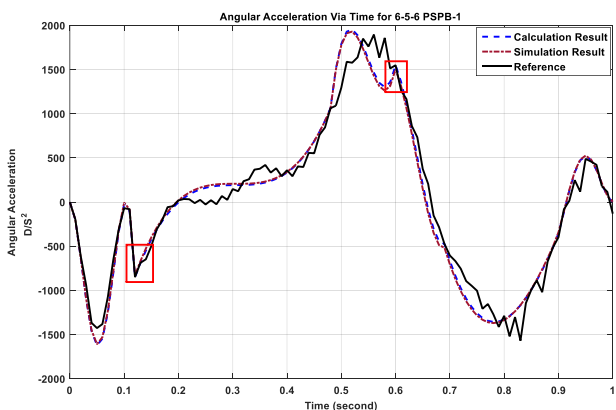 | 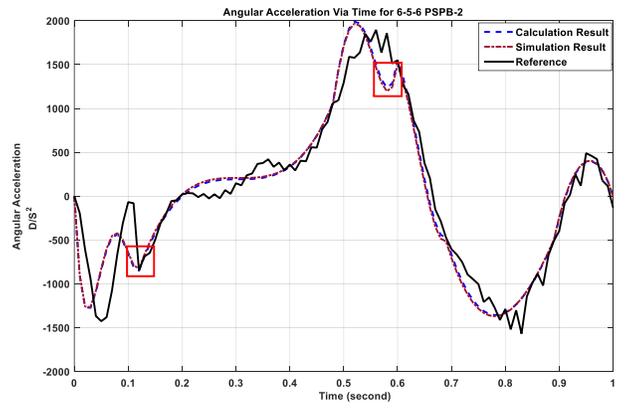 |
| | (e) | (f) |

### 3.2.3 Angular Acceleration

In term of acceleration, we can see different via points is affected directly by constraints as illustrated by red box in Table XIII.

First of all, the 4-3-4 PSPB trajectory has less effect. The reason is that, either the 4-3-4 PSPB-1 or the 4-3-4 PSPB-2 has the same constraints at the connecting points between each segment. However, the error that indicates by red box based on 4-3-4 PSPB-1, it is because of using the constraints acceleration at the initial of the first segment, which means the first segment has (P,V,A) constraints at initial and (P,V) at final point where the second segment has (P,V) at initial and (P,V) at final point and the last segment has the same constraints of the first segment. In order to proof that reason, we replace acceleration constraint by the via point position based on 4-3-4 PSPB-2. Based on that, we can notice that error terminated and became smooth. The same reason goes to the combination of $5^{th}$ and $4^{th}$ order polynomial with higher effect.

### 3.2.4 Conclusion

In conclusion, the 4-3-4 PSPB method is not sufficient to generate the trajectory for the human motion because it cannot generate the trajectory with full kinematic constraints. Furthermore, the 4-3-4 PSPB trajectory does not grantee to generate a smooth trajectory due to the linear acceleration profile during the whole mid-segment. It leads to generate infinite spike jerk during mid-segment because of the second and third derivative of the position.

Based on that problem, the 5-4-5 PSPB method is suggested to solve the linearity of the acceleration profile. However, the 5-4-5 PSPB method is not suitable to generate the trajectory because of the insufficient constraints that involved during the trajectory generation. Additionally, $4^{th}$ order polynomial has five constraints for both initial and final point where $5^{th}$ order polynomial has six constraints at initial and final points, which means, the incoordination between the constraints at either initial or final point for the first segment and the second segment, it leads to generate unexpected profile at some points.

The trajectory method of the 6-5-6 PSPB is the best among the aforementioned methods to come out with sufficient constraints. The 6-5-6 PSPB trajectory can generate the human trajectory with sufficient constraints at the initial point of the first segment and at the final point of the last segment. Also, the acceleration constraint is considered during initial and final points for all segments. Thus, the 6-5-6 PSPB trajectory method has the ability to generate a smooth profile for all angular position, velocity, acceleration, and jerk with less error.

## 3.3 Accuracy analysis at via point

For generating an accurate trajectory profile with ensuring a smooth transition through the via points, accuracy analysis is obtained.

According to a combination of polynomial equations based on PSPB technique, the 4-3-4 PSPB, 5-4-5 PSPB, and 6-5-6 PSPB are presented by considering different kinematic constraints for each PSPB technique. Furthermore, the effect of kinematic constraints on the trajectory profile is more concerned to avoid discontinuity at the via points. The ability to generate accurate trajectory profile can be evaluated based on the root mean square error (RMSE) and average difference error (ADE). For deep illustration, Figure 1, 2, and 3 shows the details of each segment and each single via point (Pv).

To ensure the smoothness of the transition from one segment to the next segment through the $P_v$, we have taken 0.01s before and after each $P_v$. Besides that, we evaluated the accuracy of each PSPB technique at each single $P_v$ for angular position, velocity, and acceleration.

### 3.3.1 Angular Position

Obviously, the effect of kinematic constraints on the trajectory profile through the via point is more clear. Based on Fig 7, the 5-4-5 PSPB-1 and 4-3-4 PSPB-1 recorded the highest RMS error among the others. We can say the reason behind that, the constraints at the end of the first segment and the constraints at the initial of the second segment are not equal or not coordinated. Furthermore, 6-5-6 PSPB-1 and 6-5-6 PSPB-2 recorded the reasonable RMS error through all the via points compared to the others that are recorded unstable RMS errors.

### 3.3.2 Angular velocity

Based on Fig 8, we noticed that the RMS error is the highest related to the 4-3-4 PSPB trajectory through most of the via points. Additionally, the 5-4-5 PSPB trajectory is randomly changed from highest to average as well as to the lowest error. From the best of my knowledge, the reason is that there is no matched between the kinematic constraints through the via point that is connected between two segments. Thus, the equality of kinematic constraints of each via point is the best way to avoid the error during generating trajectory profile.

### 3.3.3 Angular Acceleration

According to Fig 9, we noticed that the RMS error is the highest based on the 4-3-4 PSPB trajectory through most of the via points. Additionally, the 5-4-5 PSPB is randomly changed from highest to average as well as to the lowest error. From the best of my knowledge, the reason is that there is no correlation between the kinematic constraints through the via point that is connected between two segments. Thus, the equality of kinematic constraints of each via point is the best way to avoid the random changes during generating trajectory profile.

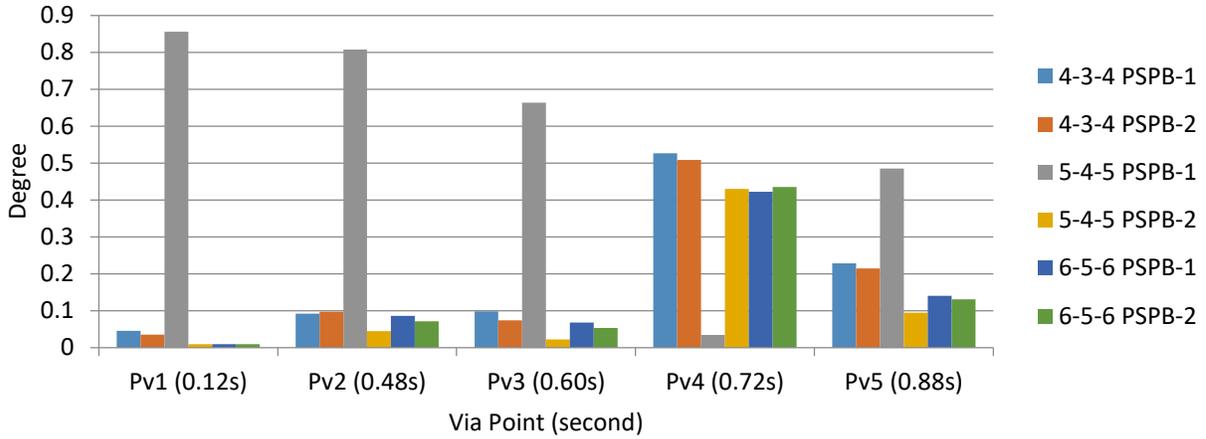

Fig. 7. Root Mean Square Error (RMSE) of Angular Position

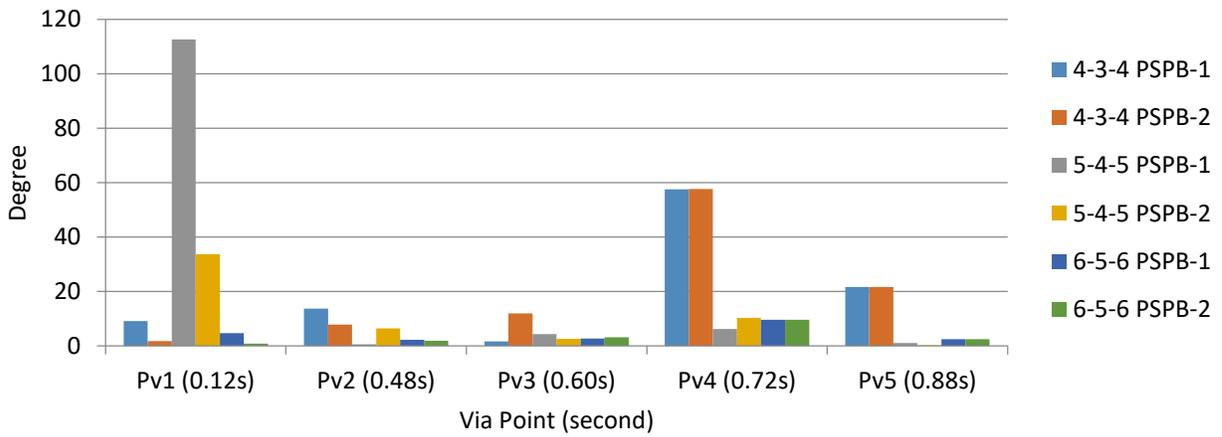

Fig. 8. Root Mean Square Error (RMSE) of Angular Velocity

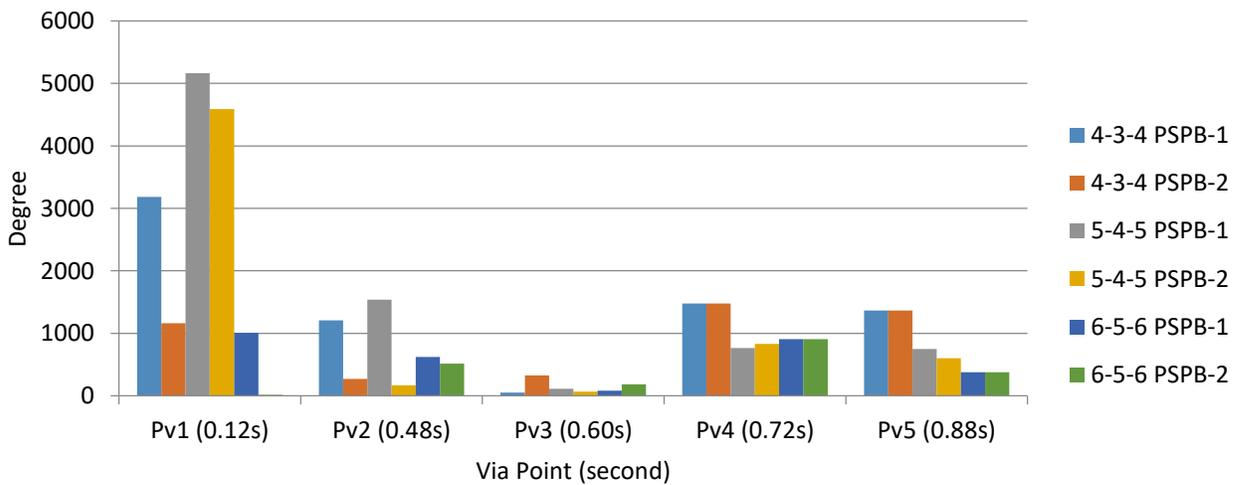

Fig. 9. Root Mean Square Error (RMSE) of Angular Acceleration

## 3.4 Executed time

In some cases, a trajectory must be modified in order to take into proper consideration to the saturation limits of the actuation system and to plan the desired motion. Thus, these limits are not violated. Motion profiles requiring values of velocity, acceleration, and torque outside the allowed ranges must be avoided since these motions cannot be performed. It is possible to see that in any case, the most restrictive constraint is due to the velocity limit and that, if the degree of the polynomial function increases, the duration (T) of the trajectory increases.

Based on running time of the algorithm, we can notice that increasing the order polynomial, the duration time of the running algorithm increases as stated in Table XIV and Fig 10.

Table XIV
Executed Time of PSPB Techniques

| PSPB Technique | Executed Time (s) |
| --- | --- |
| 4-3-4 PSPB | 0.013485 |
| 5-4-5 PSPB | 0.0141786 |
| 6-5-6 PSPB | 0.0146022 |

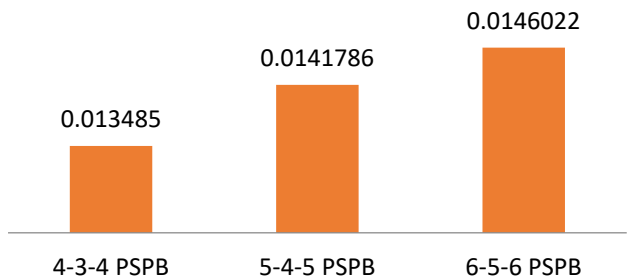

Fig. 10. Executed time

## 4. CONCLUSION

To summarize that, Trajectory planning is considered as the most concerned by researchers. Trajectory planning is a crucial part of robot design. The problem of the effect of kinematic constraints on trajectory generation directly affects the accuracy of trajectory profile. The effect of kinematic constraints at the via point is because there is no ideally match between the final point of the first segment and initial point of the next segment. Based on that, the result shows that RMSE of angular position for 4-3-4 PSPB-1, 4-3-4 PSPB-2, 5-4-5 PSPB-1, 5-4-5 PSPB-2, 6-5-6 PSPB-1, and 6-5-6 PSPB-2 are 0.4574, 0.0172, 10.9089, 0.1242, 0.6153, and 0.3128 degrees respectively. At the same time, Average difference errors are 0.0455, 0.0017, 1.0855, 0.0124, 0.0612, and 0.0311 degrees respectively. Thus, the error is increased obviously when there is no match at the via point in terms of the number of kinematic constraints.

ACKNOWLEDGEMENT

Authors would like to greatly express their thanks and appreciation to UTeM Zamalah Scheme and UTeM Centre for Research and Innovation Management (CRIM) for their encouragement, help and financially supporting to complete this research work.. This project was conducted in Center of Excellence in Robotics and Industrial Automation (CERIA) in UniversitiTeknikal Malaysia Melaka.